\documentclass[journal]{IEEEtranTIE}
\usepackage{graphicx}
\usepackage{cite}
\usepackage{picinpar}
\usepackage{amsmath}
\usepackage{url}
\usepackage{flushend}
\usepackage{colortbl}
\usepackage{soul}
\usepackage{multirow}
\usepackage{pifont}
\usepackage{color}
\usepackage[table,xcdraw]{xcolor}
\usepackage{xcolor}
\usepackage{alltt}
\usepackage[hidelinks]{hyperref}
\usepackage{enumerate}
\usepackage{siunitx}
\usepackage{breakurl}
\usepackage{epstopdf}
\usepackage{pbox}
\usepackage{amsmath,amsfonts}
\usepackage{algorithmic}
\usepackage{algorithm}
\usepackage{array}
\usepackage[caption=false,font=normalsize,labelfont=sf,textfont=sf]{subfig}
\usepackage{textcomp}
\usepackage{stfloats}
\usepackage{url}
\usepackage{verbatim}
\usepackage{graphicx}
\usepackage{cite}
\usepackage{breqn}
\usepackage{makecell}
\usepackage{tabularx}
\usepackage{adjustbox}
\usepackage{amsmath,amsfonts}
\usepackage{algorithmic}
\usepackage{algorithm}
\usepackage{array}
\usepackage[caption=false,font=normalsize,labelfont=sf,textfont=sf]{subfig}
\usepackage{textcomp}
\usepackage{stfloats}
\usepackage{url}
\usepackage{verbatim}
\usepackage{graphicx}
\usepackage{cite}
\usepackage{graphicx}
\usepackage{cite}
\usepackage{picinpar}
\usepackage{amsmath}
\usepackage{url}
\usepackage{flushend}
\usepackage{colortbl}
\usepackage{soul}
\usepackage{multirow}
\usepackage{pifont}
\usepackage{color}
\usepackage[table,xcdraw]{xcolor}
\usepackage{xcolor}
\usepackage{alltt}
\usepackage[hidelinks]{hyperref}
\usepackage{enumerate}
\usepackage{siunitx}
\usepackage{breakurl}
\usepackage{epstopdf}
\usepackage{pbox}
\usepackage{amsmath,amsfonts}
\usepackage{algorithmic}
\usepackage{algorithm}
\usepackage{array}
\usepackage[caption=false,font=normalsize,labelfont=sf,textfont=sf]{subfig}
\usepackage{textcomp}
\usepackage{stfloats}
\usepackage{url}
\usepackage{verbatim}
\usepackage{graphicx}
\usepackage{cite}
\usepackage{breqn}
\usepackage{makecell}
\usepackage[utf8]{inputenc} % UTF-8 인코딩 사용
\usepackage{tabularx}
\begin{document}
\title{A Novel 6-axis Force/Torque Sensor Using Inductance Sensors}

\author{                                                                      
	\vskip 1em
	
	Hyun-Bin Kim and Kyung-Soo Kim,~\IEEEmembership{Member,~IEEE,}

	\thanks{
	 Manuscript created January, 2025; This research was developed by the MSC (Mechatronics, Systems and Control) lab in the KAIST(Korea Advanced Institute of Science and Technology which is in the Daehak-Ro 291, Daejeon, South Korea(e-mail: youfree22@kaist.ac.kr; kyungsookim@kaist.ac.kr). (Corresponding author: Kyung-Soo Kim).  
	}
}

\maketitle
	
\begin{abstract}
This paper presents a novel six-axis force/torque (F/T) sensor based on inductive sensing technology. Unlike conventional strain gauge-based sensors that require direct contact and external amplification, the proposed sensor utilizes non-contact inductive measurements to estimate force via displacement of a conductive target. A compact, fully integrated architecture is achieved by incorporating a CAN-FD based signal processing module directly onto the PCB, enabling high-speed data acquisition at up to 4~kHz without external DAQ systems. The sensing mechanism is modeled and calibrated through a rational function fitting approach, which demonstrated superior performance in terms of root mean square error (RMSE), coefficient of determination ($R^2$), and linearity error compared to other nonlinear models. Static and repeatability experiments validate the sensor’s accuracy, achieving a resolution of 0.03~N and quantization levels exceeding 55,000 steps, surpassing that of commercial sensors. The sensor also exhibits low crosstalk, high sensitivity, and robust noise characteristics. Its performance and structure make it suitable for precision robotic applications, especially in scenarios where compactness, non-contact operation, and integrated processing are essential.

% This study proposes a novel six-axis force/torque(F/T) sensor based on inductance sensing. While conventional six-axis F/T sensors primarily utilize strain-gauges, the proposed sensor employs a non-contact mechanism that leverages changes in inductance caused by variations in the distance to a metal object. In the sensor design process, the inductance was calculated based on the coil geometry and used to guide the fabrication. The fabricated sensor was evaluated through comparative experiments with commercially available sensors, demonstrating higher sensitivity. In particular, the proposed sensor achieved more than twice the resolution of strain-gauge-based sensors of comparable size, realizing a resolution of approximately 0.03 N. Furthermore, a CAN communication-based signal processing unit was integrated within the sensor, achieving a compact design.
\end{abstract}


\markboth{arXiv}%
{}

\definecolor{limegreen}{rgb}{0.2, 0.8, 0.2}
\definecolor{forestgreen}{rgb}{0.13, 0.55, 0.13}
\definecolor{greenhtml}{rgb}{0.0, 0.5, 0.0}

\section{Introduction}
\IEEEPARstart{R}{ecent} years have witnessed rapid advancements in various robotics domains-such as collaborative robots (co-bots), quadruped robots, and humanoids~\cite{di2018dynamic,choi2023learning,kang2023view,valsecchi2020quadrupedal,liu2025fusion,chen2025adaptive}. Among these, collaborative robots (co-bots) in particular require precise force control to ensure safe and reliable interaction with human operators, leading to a sharp increase in the demand for six-axis force/torque (F/T) sensors. Currently, the majority of commercially available six-axis sensors are based on strain-gauge technology, which is widely adopted due to its proven reliability and broad applicability. However, strain-gauge-based sensors~\cite{ati_website,kim2019multi,billeschou2021low} are inherently limited by their high production costs and finite durability. These sensors typically operate via direct contact, using adhesives such as silicone to bond strain-gauges onto elastomeric substrates. This structure is vulnerable to degradation under repeated loading and external impacts, raising concerns about long-term reliability and durability. In addition, these sensors often require dedicated data acquisition (DAQ) equipment. Consequently, when multiple sensors are employed, multiple DAQ units are also needed, which poses practical limitations in the development of complex robotic systems such as multi-jointed robots.

% Among these, co-bots in particular require precise force control to ensure safe and reliable interaction with human operators, leading to a sharp increase in the demand for six-axis force/torque(F/T) sensors. Currently, the majority of commercially available six-axis sensors are based on strain gauge technology, which is widely adopted due to its proven reliability and broad applicability. However, strain gauge-based sensors are inherently limited by their high production costs and finite durability. These sensors typically operate via direct contact, using adhesives such as silicone to attach strain gauges to elastomeric substrates. This structure is susceptible to degradation under repeated loading and external impact, resulting in long-term reliability and durability concerns.
% 또한, 추가적으로 data acquisition(DAQ)장비가 필요하기 때문에 센서 여러개를 사용하기 위해서는 여러개의 DAQ장비가 필요하다. 이러한 것은 다관절 로봇과 같은 로봇 제작에 한계점이 된다. 

To address these limitations in cost and robustness, non-contact sensing technologies for F/T measurement have garnered increasing attention. A representative example is the capacitive sensing approach, which indirectly estimates applied force by monitoring changes in capacitance between electrodes and a grounded plate~\cite{kim2016novel,pu2021modeling,luo2022parameters}. Capacitive sensors are valued for their low production costs and structural simplicity, making them suitable for applications such as robotic hands and compact surgical F/T sensors~\cite{kim2017surgical,kim2018sensorized}. Their compact form factor and flexibility in design further enhance their applicability in embedded robotic systems.

In addition, optical sensing techniques have also been explored extensively~\cite{palli2013optical,al2018design,tar2011development}. These include distance measurement via photocouplers~\cite{kim2024compact,jeong2020miniature,jeong2018design} and strain detection using fiber Bragg grating (FBG) sensors~\cite{xiong2020six}. Photocoupler-based systems offer advantages in miniaturization and circuit simplicity, resulting in reduced part count and lower manufacturing cost. Conversely, FBG-based sensors provide exceptional precision and sensitivity, but require dedicated optical transducers and analysis equipment, which increase the system’s bulk and complexity. Moreover, optical sensors are generally more susceptible to environmental temperature variations compared to capacitive approaches, necessitating additional compensation mechanisms~\cite{wang2021improved}.

While these non-contact technologies each have unique advantages, they still face challenges related to cost-effectiveness, durability, compact integration, and temperature stability. In response, this study proposes a novel six-axis F/T sensor that leverages inductance variation for force detection. Inspired by the operational principle of metal detectors, the proposed sensor uses changes in inductance-caused by the displacement of nearby metallic elements within a magnetic field generated by coils-to precisely infer force. The inductive sensing approach offers robust resistance to mechanical impact, stable performance under temperature variation, and the advantage of non-contact operation. In terms of temperature sensitivity, capacitive sensors are significantly affected due to the strong dependence of permittivity on temperature. In contrast, although inductive sensors can exhibit variations in magnetic permeability with respect to temperature-particularly in magnetic materials-these changes are generally smaller and more stable, resulting in relatively lower temperature sensitivity. 

This paper systematically presents the design methodology, fabrication process, performance evaluation, and comparative analysis of the proposed inductive sensor against existing commercial alternatives.

\subsection{Contribution}

The key contributions of this study are summarized as follows:

\begin{itemize}
    \item \textbf{Introduction of a non-contact inductive sensing method}:  
    A novel measurement principle based on inductance variation is proposed to overcome the limitations of conventional strain-gauge and capacitive sensors. The proposed approach simultaneously achieves cost-efficiency and enhanced durability while offering improved resilience to environmental disturbances.

    \item \textbf{Fully integrated Printed Circuit Board(PCB)-based sensor architecture}:  
    The sensor is designed using a PCB-only structure, eliminating the need for complex mechanical assembly or additional discrete components. This significantly simplifies the manufacturing process and enhances consistency and reliability in large-scale production.

    \item \textbf{High-resolution force detection performance}:  
    By precisely modeling and calibrating the coil inductance, the proposed sensor achieves a resolution more than twice that of conventional strain-gauge sensors, reaching approximately 0.03 N. Experimental validation confirms its applicability in precision-sensitive robotic and industrial tasks.

    \item \textbf{Compact integration of onboard signal processing}:  
    In contrast to traditional sensors that rely on external signal processing units, the proposed design integrates a CAN-based signal processing module directly into the PCB. This compact integration eliminates the need for external DAQ hardware, thereby reducing wiring complexity and enhancing overall system flexibility and deployability.

    % In contrast to traditional sensors that rely on external signal processing units, the proposed design incorporates a CAN-based signal processing module directly into the PCB. This compact integration reduces wiring complexity and enhances the overall system flexibility and deployability. 외부 DAQ가 필요없다.
\end{itemize}

\section{Design of the Proposed Sensor}

\subsection{Principle of the Proposed Sensor}
 \begin{figure}[t!]
    \centerline{\includegraphics[width=\columnwidth]{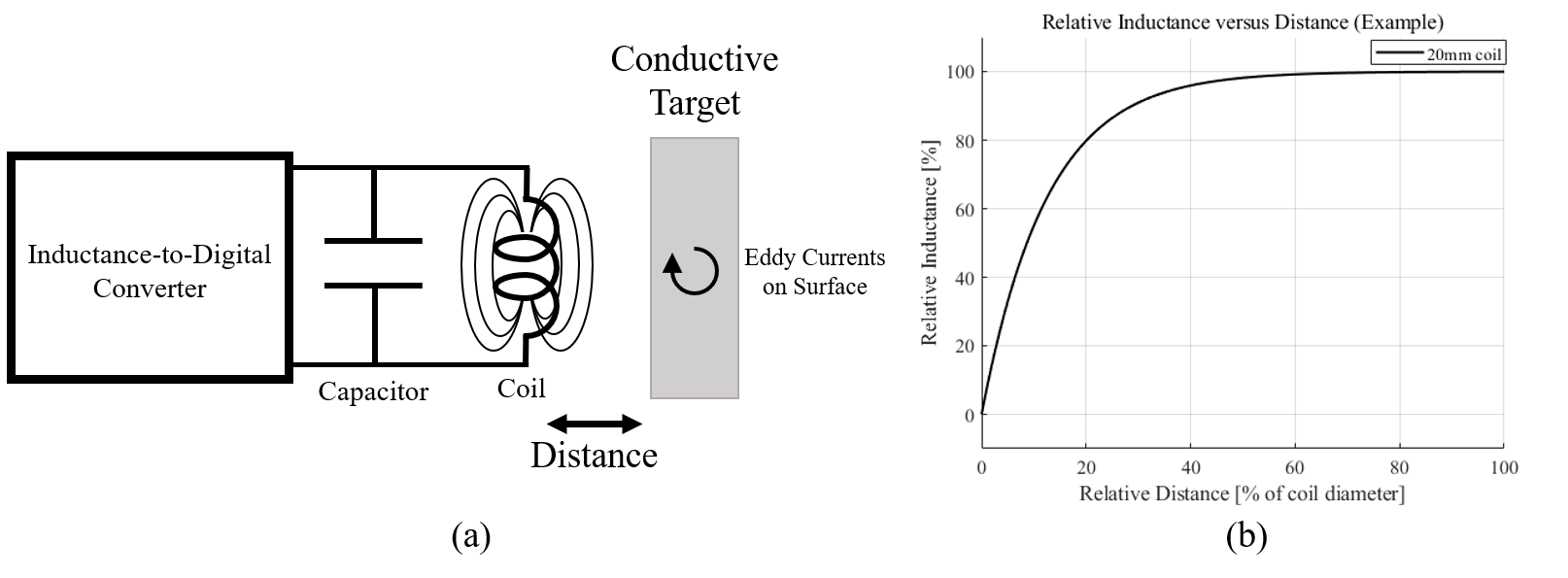}}
    \caption[Principle of the Proposed Sensor]{Principle of the proposed sensor (a). Inductance-to-digital converter with conductive target (b). Relative Inductance versus distance between coil and conductive target
    } \label{proposedprinciple}
\end{figure}
The operating principle of the proposed sensor is fundamentally based on the same principle as that of a metal detector. As illustrated in Fig.\ref{proposedprinciple}(a), the sensor measures the resonant frequency of a system composed of a capacitor and a coil, where the resonant frequency is given by $f_0=1/{2\pi \sqrt{LC}}$. When a metallic object approaches the vicinity of the inductor, the surrounding magnetic field is disturbed, leading to eddy current losses. As a result, the inductance $L$ decreases. An inductance-to-digital converter (LDC) detects this change in frequency with high precision and digitizes the measurement for output. As shown in Fig.\ref{proposedprinciple}(b), when the distance between the coil and the metallic object reaches a certain range relative to the coil diameter, the relative inductance changes noticeably. This principle enables the estimation of the distance between the coil and the metallic object. By exploiting this phenomenon, when a force is applied to an elastomer, causing a slight deformation, the resulting distance change can be detected and subsequently back-calculated to estimate the applied force.
%제안하는 센서의 원리는 기본적으로 금속탐지기의 원리와 같은 원리로 Fig.~\ref{proposedprinciple} (a)와 같이 capacitor와 coil을 이용하여 공진 주파수를 측정하는데 주파수는 $f_0=1/{2\pi \sqrt{LC}}$로 정해진다. 여기서 금속 물체가 인덕터 주변에 접근하면 인덕터 주위의 자기장이 변하면서 eddy current loss가 발생한다. 그 결과로 인덕턴스 $L$이 감소한다. 이때, inductance to digital converter가 주파수 변화를 감지하고 정밀하게 측정하여 디지털화하여 출력을 보여준다. Fig.~\ref{proposedprinciple} (b)처럼 coil의 diameter의 어느정도 거리가 있게 되면 relatative inductance가 변화하는 것을 볼 수 있다. 이처럼 코일의 반경내에 있는 금속 물체와의 거리를 판단 할 수 있다. 이를 이용하여 힘이 Elastomer에 가해지면 미세한 elastomer의 거리변화가 생기기 때문에 그것을 측정해 힘으로 역계산이 될 수 있다.   

\begin{equation}
D_{\text{in}} = D_{\text{out}} - (2N + 1)s - (2N - 1)w
\end{equation}
 $D_{\text{in}}$ denotes the inner diameter of the coil, while $D_{\text{out}}$ represents the outer diameter. Here, $N$ indicates the number of turns, $s$ is the spacing between adjacent traces, and $w$ denotes the width of a single trace.

%먼저 여기에서 D_{\text{in}}은 코일의 내경을 나타내고 D_{\text{out}}은 코일의 바깥 지름을 나타낸다. 여기서 $N$은 코일의 감은 횟수를 나타내며, $s$는 감는 간격 trace와 trace간의 spacing을 나타내고, $w$는 코일 한 줄의 두께들 나타낸다.   
\begin{equation}
D_{\text{avg}} = \left(1 + \frac{4}{\pi} \left(\frac{d_L}{D_{\text{out}}} - 1\right)\right) \cdot \frac{D_{\text{in}} + D_{\text{out}}}{2}
\end{equation}

The average diameter $D_{\text{avg}}$ is calculated using the above expression, where $d_L$ denotes the diameter measured at the center of the outermost trace. Instead of a simple arithmetic mean, a correction factor $\left(1 + \frac{4}{\pi} \left(\frac{d_L}{D_{\text{out}}} - 1\right)\right)$ is introduced to account for the physical distribution of the coil windings.

%그래서 평균 직경 계산을 위의 식과 같이 할 수 있으며, 여기서 $d_L$은 최외곽 도선의 중심을 기준으로 한 직경을 나타낸다. 여기서 내경과 외경의 평균을 내는데 단순한 평균이 아니라 보정항인 $(\left(1 + \frac{4}{\pi} \left(\frac{d_L}{D_{\text{out}}} - 1\right)\right))$을 이용해서 실제 물리적 분포를 고려한다.

\begin{equation}
\alpha = \frac{D_{\text{out}} - D_{\text{in}}}{D_{\text{out}} + D_{\text{in}}}
\end{equation}
%여기서 $\alpha$는 코일의 두께 대비 전체 크기 비율을 나타낸다. 이 값은 코일이 얼마나 두껍게 감겨있는지를 나타내고 인덕턴스 계산에 영향을 준다.
 The geometric parameter $\alpha$ quantifies the ratio of the coil thickness to its overall size. This parameter reflects how thickly the coil is wound and influences the inductance calculation.

\begin{equation}
L_{\text{layer}} = \frac{\mu_0}{2} \cdot N^2 \cdot D_{\text{avg}} \cdot \left[\ln\left(\frac{2.46}{\alpha}\right) + 0.2 \alpha^2\right]
\end{equation}

Since the PCB typically consists of multiple layers, the self-inductance $L_{\text{layer}}$ of a single layer is computed. It is determined by the permeability of free space $\mu_0$, the number of turns $N$, and the average diameter $D_{\text{avg}}$. The geometric parameter $\alpha$ is incorporated through a logarithmic and polynomial adjustment to more accurately reflect the coil's geometry.

% PCB에는 여러 Layer가 있기 때문에 한층당 자체 인덕턴스를 구할 수 있다. 각 layer의 인덕턴스인 $L_{\text{layer}}$는 permeability of free space인 $\mu_0$와 코일의 감겨있는 turn 수와 코일의 평균 직경을 이용하여 구할 수 있다. 여기서 앞에서 구한 $\alpha$를 로그 함수와 다항식을 이용하여 인덕턴스를 조정하여 한 레이어의 인덕턴스를 구할 수 있다. 

\begin{equation}
k(h) = \frac{1}{0.184 h^3 - 0.525 h^2 + 1.038 h + 1.001}
\end{equation}
Here, $h$ denotes the normalized distance between layers. The coupling factor $k(h)$ characterizes the magnetic coupling strength between adjacent layers and is essential for calculating mutual inductance between different layers.

%여기서 $h$는 layer 간 거리를 나타낸다. layer끼리 서로 자기장을 공유하면서 결합이 발생하는데 그 세기를 나타낸다. $k(h)$는 서로 다른 두 layer 사이의 인덕턴스 상호작용 정도를 계산할 때 사용된다.  

\begin{equation}
L_{\text{total}} = \left(2 \sum_{i=1}^{M-1} k_i + M\right) \cdot L_{\text{layer}}
\end{equation}
The total inductance $L_{\text{total}}$ in the absence of a metallic target is computed by combining the self-inductance and mutual inductance contributions. $M$ represents the number of layers, with $M \cdot L_{\text{layer}}$ accounting for the self-inductance and the summation term capturing the interlayer coupling.

%metal target이 없을 때 코일의 전체 인덕턴스를 구하기 위해서 자가 인덕턴스와 결합 인덕턴스를 이용하여 구할 수 있다. $M$은 layer 개수이며, $M\cdot L_{\text{layer}}$는 자가 인덕턴스 항을 나타내고 다른 항은 결합 인덕턴스를 나타낸다. 

\begin{equation}
f_{\text{res}} = \frac{1}{2\pi \sqrt{L_{\text{total}} (C + C_{\text{par}})}}
\end{equation}
The resonant frequency $f_{\text{res}}$ of the LC circuit without a metallic target is calculated using the total inductance $L_{\text{total}}$ and the capacitance $C$ combined with the parasitic capacitance $C_{\text{par}}$. In this study, a parasitic capacitance of 4 pF is assumed based on the circuit layout.

% metal target이 없을 때 코일로 인한 LC회로에서 공진 주파수를 구할 수 있으며, 여기서 $C_{\text{par}}$는 parasitic capacitance로 일반적으로 회로 구성으로 생기는 것을 나타내고 본 논문에서 4 pF의 값을 사용하였다.

% \begin{equation}
% \Delta L = L_{\text{total}} - L' = L_{\text{layer}} \cdot \left[ \left(2 \sum k_i^{\text{target}} + M \right) - \left(2 \sum k_i + M \right) \right]
% \end{equation}
\begin{equation}
\resizebox{.95\linewidth}{!}{$
\Delta L = L_{\text{total}} - L' = L_{\text{layer}} \cdot \left[ \left(2 \sum k_i^{\text{target}} + M \right) - \left(2 \sum k_i + M \right) \right]
$}
\end{equation}
When a conductive target interacts with the coil, the inductance decreases. The proximity of the target modifies the coupling coefficients $k_i^{\text{target}}$ between layers, leading to a change in total inductance $\Delta L$. This variation can be correlated to external force applied to the elastomer.~\cite{grover2004inductance,mohan1999simple,ti_website}

The LDC measures the resonant frequency $f_{\text{res}}' = \frac{1}{2\pi \sqrt{L_{\text{total}}' (C + C_{\text{par}})}}$ to detect changes in inductance. As the LDC measures a value inversely proportional to the frequency, the raw digital output is effectively proportional to $\sqrt{L}$.

%metal target과 상호작용을 할 때 인덕턴스 값은 줄어들게 되는데 conductive target이 가까이 왔을 때 각 layer 간 결합계수$k_i^{\text{target}}$가 달라지는데 이를 이용하여 전체 인덕턴스가 변해서 $\Delta L$을 구살 수 있어 힘 변화를 유추할 수 있다.
% inductance to digital converter는 frequency 즉 공진 주파수를 측정하여 인덕턴스의 변화를 알아내기 때문에 $f_{\text{res}} = \frac{1}{2\pi \sqrt{L_{\text{total}^'} (C + C_{\text{par}})}}$를 이용하여 유추할 수 있다. 
% inductance to digital converter는 frequency에 반비례 하는 형태로 측정을 하게 되는데 출력 Raw data는 $\sqrt{L}$에 비례하게 된다. 

\subsection{Elastomer and Printed Circuit Board Design}

\begin{figure}[!t]\centering
	\includegraphics[width=1\columnwidth]{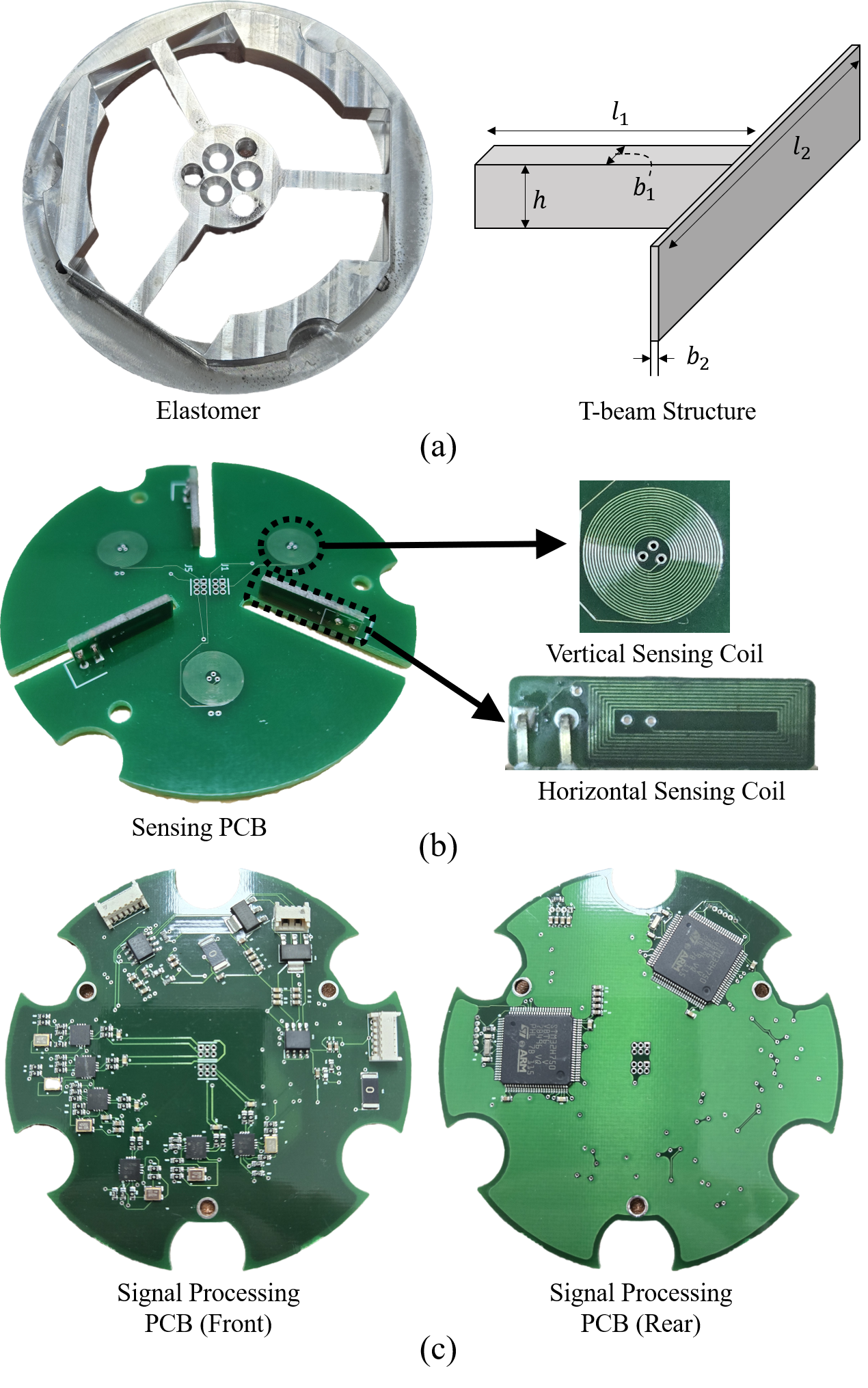}
	\caption{Elastomer and printed circuit board design (a) Elastomer and T-beam structure (b) Sensing PCB with vertical sensing coil and horizontal sensing coil (c) Signal processing PCB design
  }\label{pcbdesign}
\end{figure}

\begin{figure}[!t]\centering
	\includegraphics[width=0.8\columnwidth]{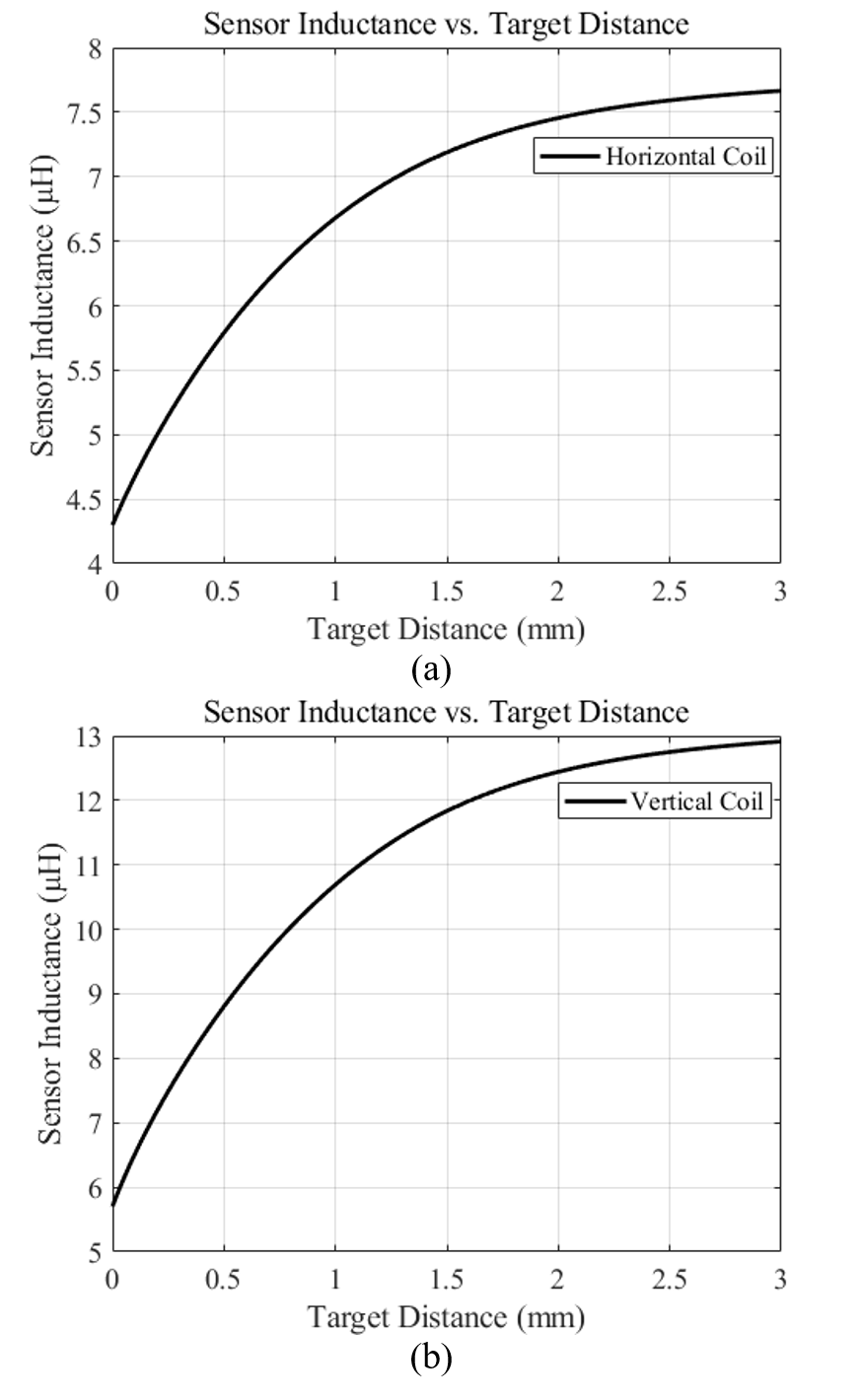}
	\caption{Calculated graph of coils' inductance and conductive target distance (a) Horizontal coil (b) Vertical coil
  }\label{target}
\end{figure}
%Elastomer는 기존에 많이 사용하는 T-beam을 이용하였다. 스트레인 게이지를 사용하는 6축 힘 토크 센서에서 가장 많이 사용하는 beam이며 여기서 Fig.~\ref{target} (a) 와 같이 Elastomer에 T-beam을 120도 간격으로 세개를 배치하였으며, 각각의 $l_1 , l_2 , b_1 , b_2 , h$는 Table. ~\ref{valtbeam}에서 나온 것과 같이 하였으며, 최소 800 N의 힘과 27 N$\cdot$m의 모멘트를 견딜 수 있도록 제작하였다.   
The elastomer structure adopts the widely used T-beam design, which is one of the most common configurations in six-axis F/T sensors employing strain-gauges. As illustrated in Fig.\ref{pcbdesign}(a), three T-beams are arranged at 120-degree intervals within the elastomer. The corresponding geometric parameters $l_1$, $l_2$, $b_1$, $b_2$, and $h$ are defined as listed in Table\ref{valtbeam}. The structure is designed to withstand a minimum of 800 N of force and 27 N$\cdot$m of torque.
\begin{table}[h!]
    \centering
    \caption{Value of T-beam}
    \begin{tabular}{cccccc}
    \hline\hline
  
         Variable & $l_1$ & $l_2$ &$b_1$ &$b_2$ & $h$   \\
         \hline
         Value (mm) & 21.5 & 31.7 & 4 & 0.5 & 7\\
    \hline
    \end{tabular}
    
    \label{valtbeam}
\end{table}
The sensing PCB of the proposed sensor, as illustrated in Fig.~\ref{pcbdesign}(b), comprises two distinct types of coils: vertical sensing coils and horizontal sensing coils. The vertical sensing coils are primarily utilized for measuring the torque components \( T_x \), \( T_y \), and the axial force \( F_z \), whereas the horizontal sensing coils are designed to detect the lateral forces \( F_x \), \( F_y \), and the torsional torque \( T_z \).

The vertical sensing coils are implemented with 18 turns distributed across three layers. Each trace has a width of 4~mil and an inter-trace spacing of 4~mil, with a copper thickness of 1~oz and a coil diameter of 10~mm. The spacing between the first and second layers is 59~mil, while that between the second and third layers is 5.9~mil.

The horizontal sensing coils consist of 10 turns across four layers, with an overall height of 4.7~mm and a width of 13~mm. The spacing between the third and fourth layers is also set to 5.9~mil.

Fig.~\ref{pcbdesign}(c) shows the signal processing printed circuit board, which incorporates the LDC1614 chip by Texas Instruments. A high-performance STM32H750 microcontroller is integrated into the board, enabling direct communication via the CAN protocol. CAN-FD communication, supporting data rates up to 5~Mbps, is utilized for high-speed data transmission.

To support high-frequency sampling, each sensing coil is connected to a dedicated LDC1614 channel, allowing for parallel and rapid inductance measurements. The system operates with a 5~V input voltage and provides four external connections: power input, ground, CAN High, and CAN Low. This architecture enables a compact and fully integrated sensor module, well-suited for deployment in embedded robotic systems.

\subsection{Sensor Fabrication}
\begin{figure*}[!th]
    \centering
    \includegraphics[width=2\columnwidth]{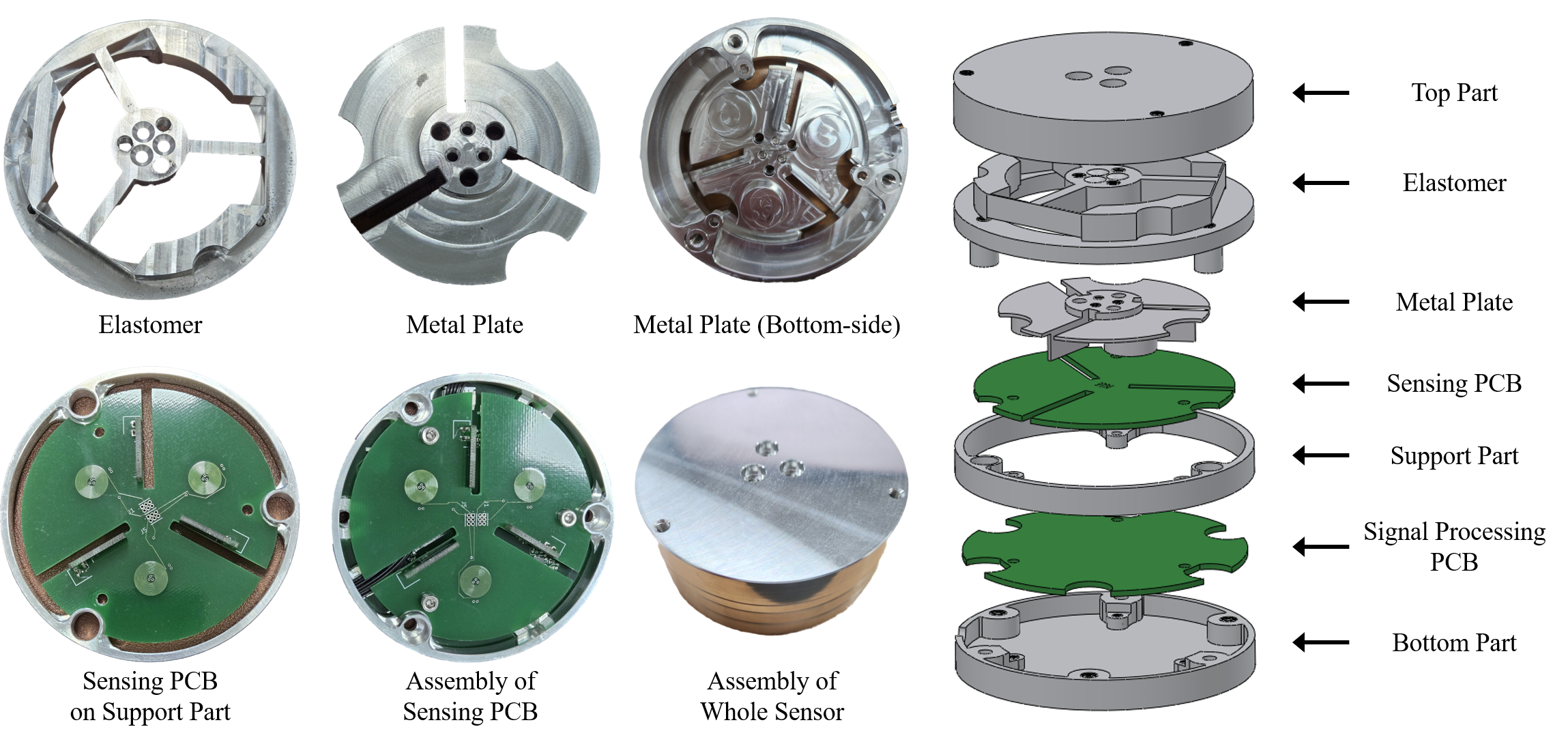} % 90도 회전 후 세로 방향 크기를 페이지에 맞춤
    \caption{Components of the proposed sensor; Five metallic components: Top part, elastomer, metal plate, support part and bottom part, two PCB components: Sensing PCB and Signal Processing PCB. }
    \label{descomp}
\end{figure*}
%센서는 총 5개의 metal 가공품과 두개의 pcb로 이루어져 있다. 여기서 가공을 이용하여 AL7075-t6 의 재질로 제작을 하였으며, Aluminium 중에서 히스테레시스가 적은 편에 속한다. 센서의 각 컴포턴트와 조립과정은 Fig.~\ref{descomp}에 나온 순서대로 되면서 bottom part에 signal processing pcb가 있고 sensing pcb를 고정해줄 bottom part와 연결이 되는 support part가 있으며, 그 위에 coil에서 거리를 측정할 metal plate가 있으며, elastomer에 달려 있다. elastomer는 support part와 bottom part에 연결이 되어 볼트로 고정된다. top part는 외부에서 힘을 가해줄 부분이다. metal plate가 구리면 감도는 더 좋지만 aluminium으로 해도 충분하기 때문에 aluminium을 사용하였다.  
The proposed sensor consists of five machined metal components and two PCBs. All metallic parts are fabricated using AL7075-T6, an aluminum alloy known for its relatively low hysteresis compared to other aluminum grades. As illustrated in Fig.~\ref{descomp}, the assembly process follows a bottom-up structure.

As shown in Fig.~\ref{descomp}, the sensor consists of five metallic components and two PCB parts. All components are fastened using bolts in combination with a threadlocker adhesive. At the base of the sensor, the signal processing PCB is enclosed within the bottom part. A support structure is mounted above the base to secure the sensing PCB in position. A metal plate is placed above the sensing PCB, serving as the inductive target for distance measurements by the coils. This metal plate is mechanically coupled to the elastomer, which is clamped between the support part and the bottom part using bolts, thereby ensuring structural integrity.

% Fig.~\ref{descomp}에 보면 5개의 metalic components랑 2개의 PCB파트가 있는것을 볼 수 있다. 모든 components들은 쓰레드락 접착제를 이용한 볼트를 이용하여 고정할 수 있다. At the base of the sensor, the signal processing PCB is housed within the bottom part. A support part is mounted above it to secure the sensing PCB in place. Positioned above the sensing PCB is a metal plate, which serves as the inductive target for distance measurements via the coils. This metal plate is mechanically attached to the elastomer, which is fixed between the support part and the bottom part using bolts, ensuring structural stability.

The top part functions as the interface through which external forces are applied to the sensor. Although copper offers higher sensitivity due to its greater conductivity, aluminum is selected for the metal plate to maintain sufficient performance while ensuring material consistency and manufacturability across the structure.

\begin{table}
\begin{center}
\caption{Mechanical specification of the prototype sensor} \label{sensrange}
\begin{tabular}{c c c c c c c}
\hline\hline
  &Value    &Unit       \\ \hline
  Diameter&85&mm\\
  Height&       36&     mm\\
Material&       AL7075-T6  & \\
Input force range ($F_x$ and $F_y$)     &$\pm$890&    N \\
Input force range ($F_z$)     &$\pm$1,435&    N \\
Input torque range ($T_x$, $T_y$) &$\pm$27        &N$\cdot$m \\
Input torque range ($T_z$)    &$\pm$45&       N$\cdot$m \\
Sampling frequency      &4&     kHz \\
Weight  &280&    g \\
 \hline
\end{tabular}
\end{center}
\end{table}

%센서의 sensor measuring range는 Table.~\ref{sensrange}에 나온 것처럼 최소 890 N의 힘과 27N$\cdot$m의 모멘트를 측정할 수 있고 최대 1435N의 힘과 45N$\cdot$m의 모멘트를 측정할 수 있다. x축과 y축의 힘은 T-beam의 구조상 거의 같으며, x축과 y축의 모멘트도 같은 것을 볼 수 있다.
Table~\ref{sensrange} summarizes the mechanical specifications of the proposed sensor. The sensor has a diameter of 85~mm and a height of 36~mm, and it operates at a sampling rate of approximately 4~kHz, which corresponds to the maximum rate supported by the LDC1614 IC (4.08~kHz). Including the PCB, the total weight of the sensor is 280~g.

As also shown in Table~\ref{sensrange}, the sensor is designed to measure forces in the range of 890~N to 1435~N, and torques ranging from 27~N$\cdot$m to 45~N$\cdot$m. Owing to the symmetric configuration of the T-beam structure, the sensor demonstrates nearly identical force responses along the $x$- and $y$-axes. Likewise, the torque responses about the $x$- and $y$-axes exhibit similar characteristics, ensuring consistent performance in planar loading conditions.

% Table.~\ref{sensrange}를 보면 센서의 mechanical specification에 대한 내용이 나온다. 센서의 diameter는 85~mm이며, 36~mm의 높이를 가지고 있으며, LDC1614 IC의 최대 샘플링 rate인 4.08~kHz로 약 4~kHz를 나타낸다. 무게는 PCB 포함 280~g의 무게를 가지고 있다. 

% As summarized in Table~\ref{sensrange}, the proposed sensor is capable of measuring forces ranging from a minimum of 890~N to a maximum of 1435~N, and moments from 27~N$\cdot$m to 45~N$\cdot$m. Due to the symmetric configuration of the T-beam structure, the sensor exhibits nearly identical performance in both the $x$- and $y$-axis force measurements. Similarly, the moment responses about the $x$- and $y$-axes are also comparable.

\section{Sensor Evaluation}
% 센서는 ATI-ia사의 MINI-85 SI-1900-80의 calibration 된 센서와 비교를 하여 평가를 진행하였다. 평가를 진행하기 전에 칼리브레이션을 진행해야 한다. 
The performance of the proposed sensor was evaluated by comparison with a calibrated commercial six-axis F/T sensor, the MINI-85 SI-1900-80 model from ATI Industrial Automation. Prior to evaluation, a calibration procedure was required.

\subsection{Calibration}
\begin{figure}[!thb]
    \centering
    \includegraphics[width=\linewidth]{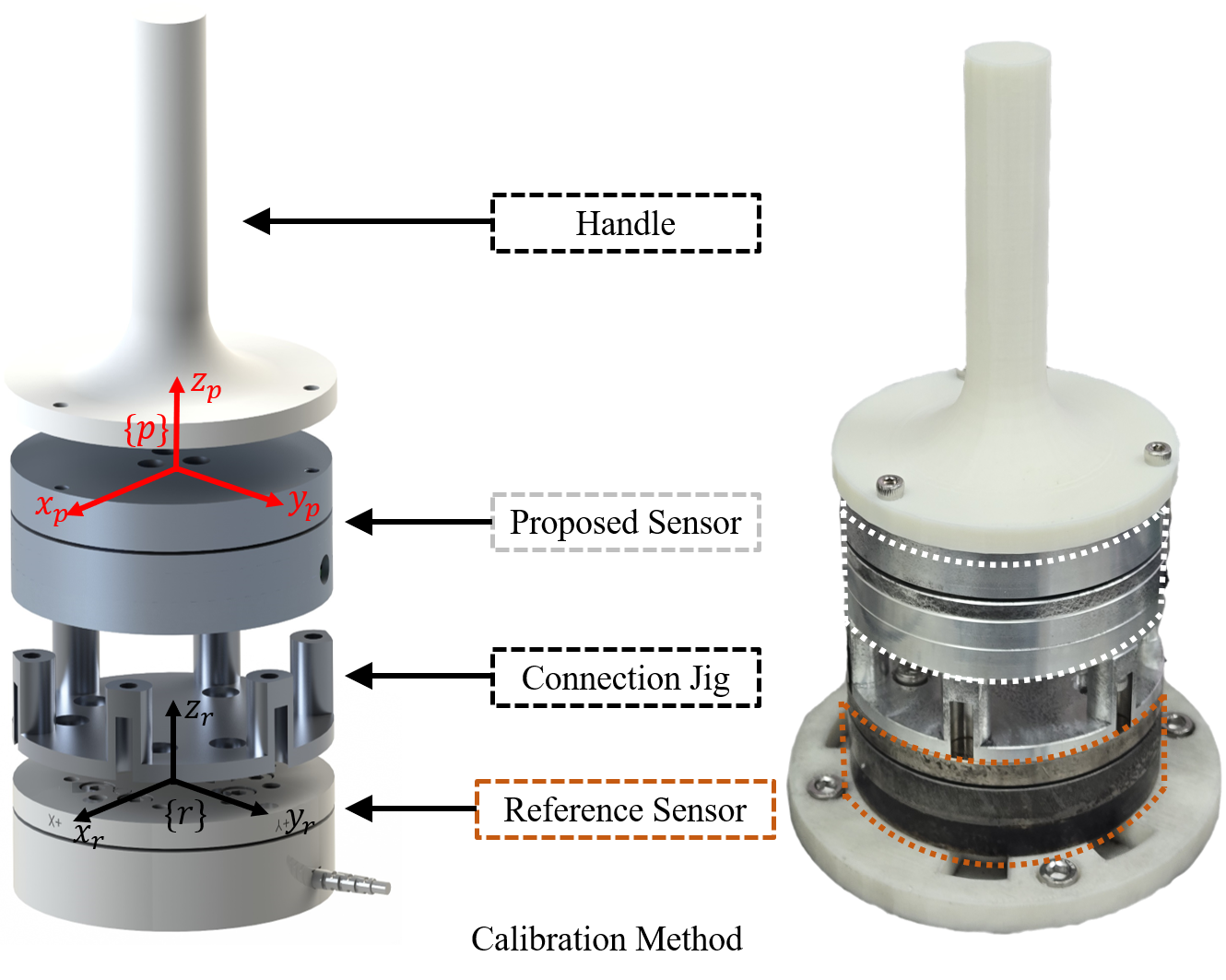}
    \caption{Calibration method with direct connecting along the $z$-axis with connection jig; F/T applied via handle for calibration.
}
    \label{cali}
\end{figure}
\begin{figure}[!thb]
    \centering
    \includegraphics[width=\linewidth]{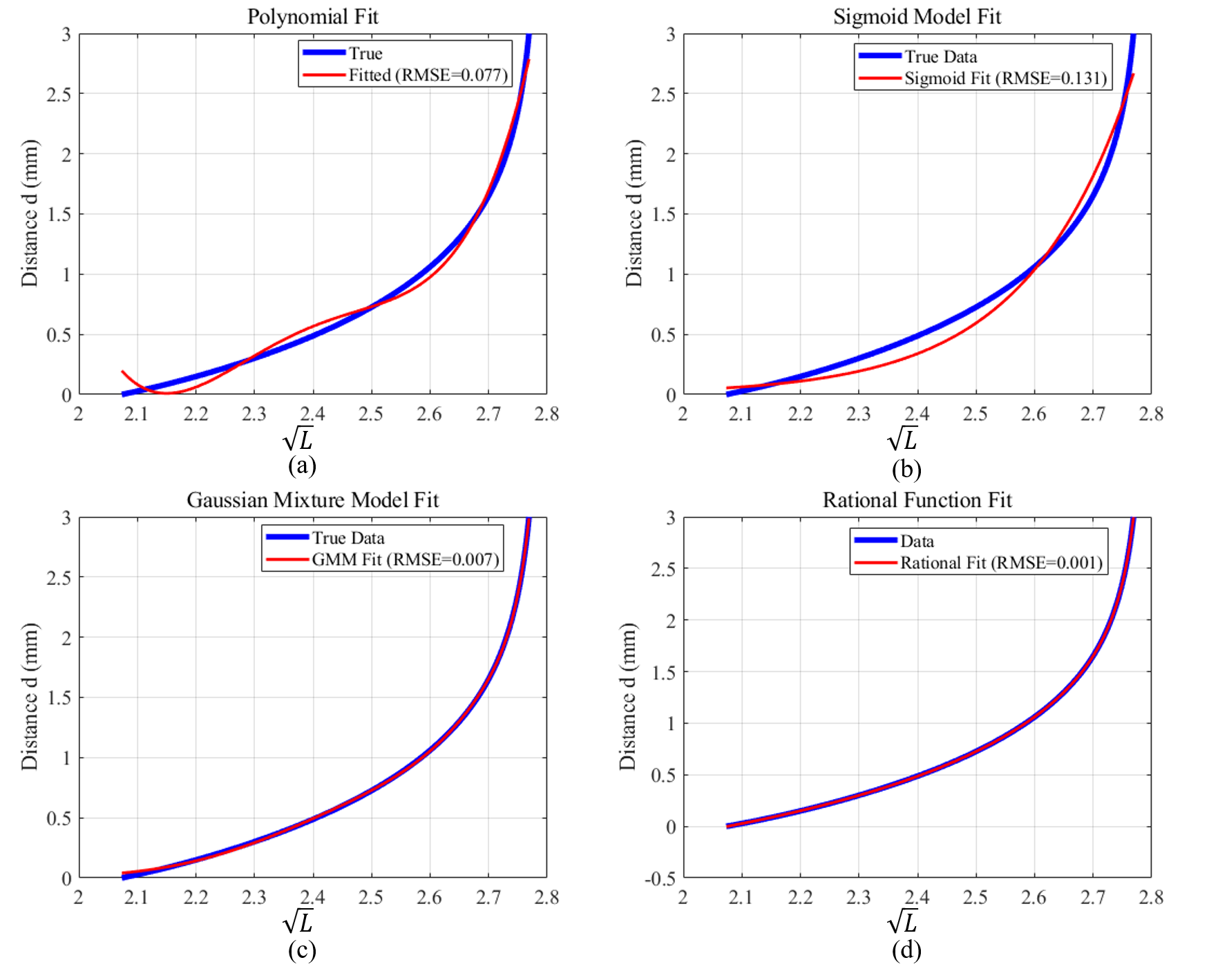}
    \caption{Fitting result of each method: (a) Polynomial fitting; (b) Sigmoid model fitting; (c) Gaussian mixture model fitting; (d) Rational function fitting.}
    \label{fitting}
\end{figure}
For calibration, a widely adopted method from the literature was employed, wherein direct force application is used to relate known inputs to sensor outputs. As illustrated in Fig.~\ref{cali}, the proposed sensor was rigidly connected to the reference sensor via a custom-designed connection jig, and external forces were applied manually using a handle.

Each sensing coil in the proposed sensor produces a digital output via an inductance-to-digital converter, which reflects the square root of the inductance value, i.e., \(\sqrt{L}\). Since the inductance \(L\) exhibits a nonlinear relationship with the distance between the sensing coil and the target metal plate, constructing a conventional linear calibration matrix results in reduced accuracy. Therefore, a more robust nonlinear calibration approach is necessary to accurately map sensor outputs to the corresponding F/T.

% 센서의 칼리브레이션은 여러 논문들에서 많이 사용하는 방법인 직결해서 힘을 측정하여 칼리브레이션 하는 방식을 사용하였다. 여기서 칼리브레이션은 칼리브레이션과 평가 장치는 Fig.~\ref{cali}와 같이 reference sensor와 proposed sensor를 connection jig로 직결하고 handle을 이용하여 힘을 주는 방식이다. 앞에서 구한 센서의 각 코일마다 inductance 값을 inductance to digital converter가 측정을 하여 나타내주는데 $\sqrt{L}$의 형태로 나타나는데 $L$이 거리에 따라서 비선형성을 나타내기 때문에 일반적인 칼리브레이션 방식인 칼리브레이션 matrix를 선형으로 구하는 것으로 하기에는 정확도가 낮다.
% 그래서 여기서는 3차 방정식을 이용하여 6*19 matrix를 least square method를 이용하여 칼리브레이션을 진행하였다. 
\begin{equation}
    F=\mathbf{A}y,\ y= \begin{bmatrix}
        y_1\\ y_2 \\y_3 \\y_4\\y_5\\ y_6  \\1
    \end{bmatrix}
\end{equation}
\begin{equation}
    y=f({x_{\text{raw}}})
\end{equation}

Here, $F$ denotes the F/T vector, and $y$ represents the deformation. The variable $x_{\text{raw}}$ refers to the raw data output from the LDC1614, and the function $f$ describes the relationship between $x_{\text{raw}}$ and $y$. Since the LDC1614 output is proportional to frequency, it is also proportional to $\sqrt{L}$. The matrix $\mathbf{A}$ is a $6 \times 7$ calibration matrix that maps the deformation to the corresponding F/T values.

% 여기서 $F$는 F/T를 나타내며, $x$는 deformation을 나타내며, $x_{\text{raw}}$과 $x$간의 관계를 나타내는 함수인 $f$로 나타나게 된다. 여기서 $x_{\text{raw}}$는 LDC1614에서 나오는 raw data를 말한다. LDC1614에서 나오는 데이터는 frequency에 대해서 비례하기 때문에 $\sqrt{L}$과 비례적인 관계로 이루어져 있다. $\textbf{A}$는 6$\times$7 matrix로 calibration matrix를 나타낸다.

%LDC1614가 측정하는 것은 $\sqrt{L}$이기 때문에 $\sqrt{L}$와 Distance 간의 fitting을 칼리브레이션에 이용하려고 하였다. Fig.~\ref{fitting}은 polynomial fitting과 sigmoid model fitting과 gauusian mixture model fitting과 rational function fitting을 비교한 것이다. 각각 모두 5개혹은 6개의 파라미터를 가지고 있는데, 각각의 식은 다음과 같다. 
Since the LDC1614 measures the square root of inductance $\sqrt{L}$, the relationship between $\sqrt{L}$ and the target distance was used for calibration. Fig.~\ref{fitting} compares the results of polynomial fitting, sigmoid model fitting, Gaussian mixture model fitting, and rational function fitting. Each model includes five or six parameters, and their respective formulations are given as follows.

\begin{equation}
    y=a_4x^4+a_3x^3+a_2x^2+a_1x+a_0
\end{equation}
%여기서 $x$는 $\sqrt{L}$을 나타내고 $y$는 distance d를 나타낸다. $a_n$는 각 $n$번째의 polynomial 계수를 나타낸다. 
Here, $x$ denotes $\sqrt{L}$ and $y$ represents the target distance $d$. The coefficients $a_n$ correspond to the $n$-th order polynomial terms. 

\begin{equation}
    y=b_1/(1+\exp{(-b_2 (x-b_3))})+b_4/(1+\exp{(-b_5 (x-b_6))})
\end{equation}
Similarly, in the sigmoid model, $x$ denotes $\sqrt{L}$ and $y$ represents the target distance $d$. The fitting result using the sigmoid function is shown in Fig.~\ref{fitting}(b), where the root mean square error (RMSE) was found to be 0.131. Each parameter $b_n$ represents the $n$-th coefficient of the sigmoid model.

% 이 식에서도 마찬가지로 $x$는 $\sqrt{L}$을 나타내고 $y$는 distance d를 나타내는데, sigmoid 함수를 이용해 fitting을 한 결과를 Fig.~\ref{fitting}(b)에서 나타난다. 여기서 RMSE는 0.131이 나왔다. 각각 $b_n$은 $n$번째 sigmoid 함수의 계수를 나타낸다. 이와 마찬가지로 nonlinear model fitting에 많이 사용하는 모델인 Gaussian Mixture Model fitting은 다음과 같이 나타낼 수 있다.
\begin{equation}
    y=c_1 \exp{(-(x-c_2)^2/(2c_3^2))}+c_4 \exp{(-(x-c_5)^2/(2c_6^2))}
\end{equation}
In the same manner, a commonly used nonlinear model fitting technique, the Gaussian mixture model, is formulated as follows. The coefficients $c_n$ denote the parameters of the Gaussian mixture model, with a total of six parameters. As shown in Fig.~\ref{fitting}(c), this model yielded a relatively lower RMSE of 0.007 compared to the previous two models.

% 여기에서 $c_n$은 $n$번째의 Gaussian Mixture Model의 계수를 나타내며, 총 6개까지 있다. 
% 여기서 RMSE는 0.007로 비교적 앞의 두개보다 낮은 수치를 Fig.~\ref{fitting}(c)에서 볼 수 있다. 
\begin{equation}
    y=(d_1 x^2+d_2 x +d_3)/(d_4x^2 +d_5 x+1)
\end{equation}
Finally, the rational function fitting model is presented. The coefficients $d_n$ represent the parameters of the rational function. As illustrated in Fig.~\ref{fitting}(d), the rational function exhibits a significantly lower RMSE of 0.001 compared to the other models. This suggests that the structural nature of $\sqrt{L}$ is particularly well-suited to rational function fitting, and thus, this model was employed for calibration. To simultaneously determine the rational function coefficients and the calibration matrix, an optimization-based approach was adopted. Specifically, MATLAB's \texttt{lsqnonlin} function was utilized for the calibration procedure.

\begin{table}[!h]
    \centering
    \caption{Comparison of each fitting function}
    \begin{tabularx}{\linewidth}{l c c c >{\centering\arraybackslash}X}
        \hline\hline
        Model & \# Parameters & RMSE & $R^2$ & Linearity Error (\%) \\\hline 
        Polynomial function & 5 & 0.0488 & 0.9968 & 4.73 \\
        Sigmoid function    & 6 & 0.1307 & 0.9774 & 11.30 \\
        Gaussian model      & 6 & 0.1404 & 0.9739 & 12.04 \\
        Rational function   & 5 & \textbf{0.0009} & \textbf{1.0000} & \textbf{0.11} \\\hline
    \end{tabularx}
    \label{fittable}
\end{table}
% \begin{table}[!h]
%     \centering
%     \caption{Comparison of each fitting function}
%     \begin{tabularx}{\linewidth}{l c c c c}
%         \hline\hline
%         Model & \# Parameters & RMSE & $\text{R}^2$ & Linearity Error (\%) \\\hline 
%         Polynomial function & 5 & 0.0488 & 0.9968 & 4.73 \\
%         Sigmoid function    & 6 & 0.1307 & 0.9774 & 11.30 \\
%         Gaussian model      & 6 & 0.1404 & 0.9739 & 12.04 \\
%         Rational function   & 5 & \textbf{0.0009} & \textbf{1.0000} & \textbf{0.11} \\\hline
%     \end{tabularx}
%     \label{fittable}
% \end{table}
To quantitatively evaluate each fitting model, three key metrics were computed: RMSE, coefficient of determination ($R^2$), and linearity error (maximum deviation as a percentage of full scale). Table~\ref{fittable} summarizes the results.

Among the compared models, the rational function achieved the lowest RMSE (0.0009), the highest $R^2$ score (1.0000), and the smallest linearity error (0.11\%). Notably, it also required only five parameters, whereas other nonlinear models, such as the sigmoid function, used six. These results suggest that the rational model not only provides the best fit but also demonstrates superior generalizability and robustness.

\subsection{Evaluation}
\begin{figure*}
    \centering
    \includegraphics[width=\linewidth]{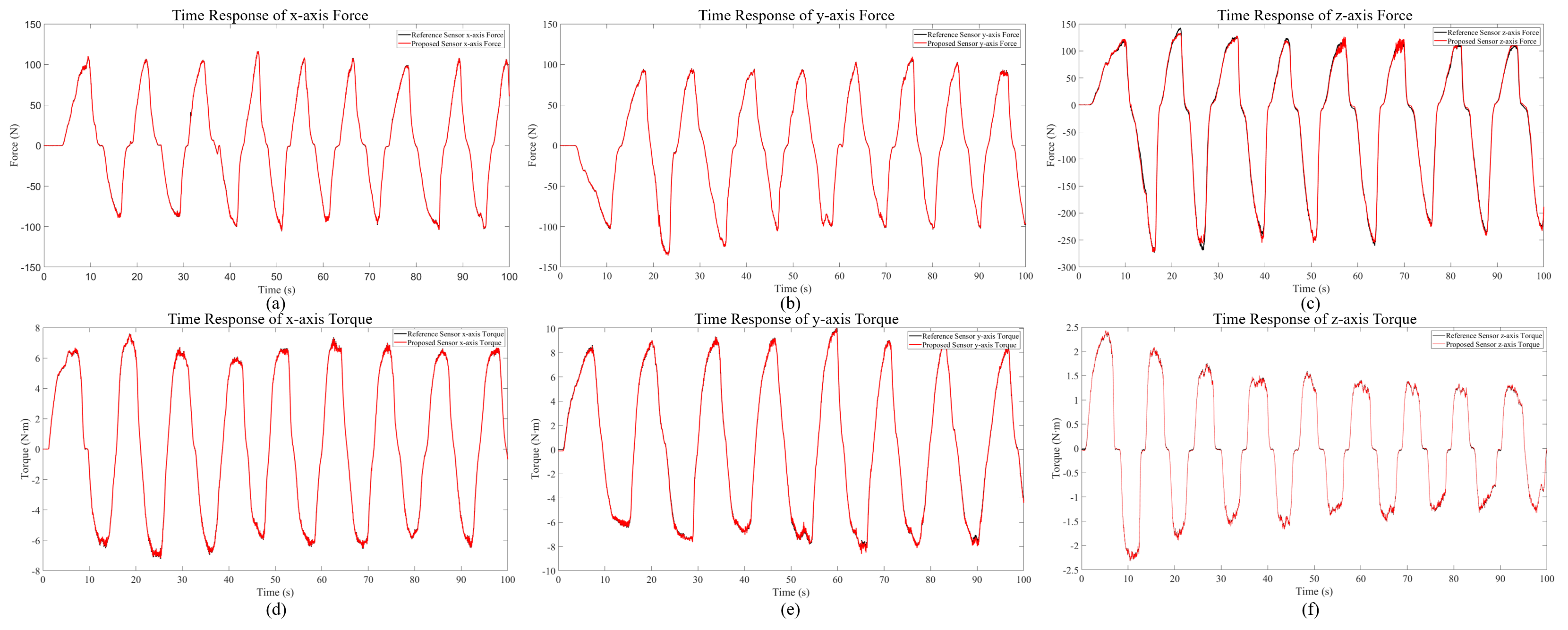}
    \caption{Static time response evaluation of each axis F/T : (a) x-axis force; (b) y-axis force; (c) z-axis force; (d) x-axis torque; (e) y-axis torque; (f) z-axis torque.}
    \label{static}
\end{figure*}
%평가는 기존의 논문들에서 많이 사용하는 static force를 측정하고 time response에 대한 rmse와 percentage error를 측정하였다. percentage error는 full-scale error로 reference sensor와 proposed sensor의 차이를 proposed sensor의 range로 나눈 값이며 이를 이용하여 계산을 하였다. 측정은 각 축 마다 100초 씩 진행을 하였다. 
%Fig.~\ref{static}은 각 축의 결과를 보여준다. reference sensor와 proposed sensor간 차이는 거의 없는 것을 볼 수 있다. 여기서 reference sensor와 proposed sensor는 둘다 1kHz로 sampling을 하였으며, reference sensor는 838Hz의 low pass filter를 이용하였다. 실험은 reference sensor는 ethernet udp 통신을 이용하여 데이터를 받고 proposed sensor는 CAN 통신을 사용하였으며, 실험은 simulink desktop realtime에서 진행하였다. 

% Table.~\ref{erroranal}은 proposed sensor의 error를 나타낸 표이며, 여기서 percentage error의 최대 크기는 힘에서는 z축 힘이 0.64\% 로 나타났으며, y축 모멘트가 0.75\%로 나타났다. 여기서 상용 센서들의 기준들은 보통 각 축의 percentage error를 1\% 로 나타내기 때문에 상용센서의 기준을 맞춰있다고 볼 수 있다. 

The sensor performance was evaluated using a static force measurement approach, which is commonly adopted in previous studies~\cite{kim2024compact}. Two metrics were used to quantify the performance: RMSE of the time response, and percentage error. The percentage error was computed as the full-scale error, defined as the difference between the proposed sensor and the reference sensor divided by the full measurement range of the proposed sensor.

Each measurement was conducted for 100 seconds per axis. During this period, forces were repeatedly applied multiple times along each axis, effectively serving as a repeatability test in addition to static evaluation. 

Fig.~\ref{static} presents the results for all six axes, showing that the proposed sensor exhibits a close match with the reference sensor across all directions. Both sensors were sampled at 1~kHz, and the reference sensor employed an 838~Hz low-pass filter. During the experiments, the reference sensor transmitted data via Ethernet UDP communication, while the proposed sensor utilized CAN communication. All experiments were conducted using Simulink Desktop Real-Time. Notably, the proposed sensor did not apply any low-pass filtering during the measurements.

\begin{table}[!h]
\centering
\caption{Full-Scale Error and RMSE analysis}
\begin{tabular}{cccccc}
\hline\hline
     & \multicolumn{3}{c}{Percentage Error (\%)} & \multicolumn{1}{c}{RMS  Error}  \\ 
     & \multicolumn{1}{c}{Mean}    & \multicolumn{1}{c}{Std}    & Max    & \multicolumn{1}{c}{(N, N$\cdot$m)}  \\ \hline
$F_x$ & 0.0119 & 0.0154 & 0.151 & 0.310  \\ 
$F_y$ & 0.0125 & 0.0161 & 0.0916 & 0.318   \\ 
$F_z$ & 0.102 & 0.137 & 0.641 & 3.92   \\ 
$T_x$ & 0.0346 & 0.0452 & 0.209 & 0.0244 \\ 
$T_y$ & 0.0891 & 0.120 & 0.749 & 0.0648   \\ 
$T_z$ & 0.00887 & 0.0110 & 0.0728 & 0.00989 \\ \hline
\end{tabular}
\label{erroranal}
\end{table}

Table~\ref{erroranal} summarizes the error analysis of the proposed sensor. The maximum percentage error observed in force measurements was 0.64\% in the $F_z$ direction, while the highest error in torque measurement was 0.75\% in the $M_y$ direction. Since most commercial six-axis F/T sensors specify a full-scale percentage error of approximately 1\% per axis, the proposed sensor can be considered to meet the standard performance requirements of commercial-grade systems.

%또한, 센서의 평가를 위해서 resolution 평가를 진행하였다. resolution은 proposed sensor와 elastomer 구조가 비슷하면서 제일 많이 사용하는 strain-gauge를 사용하는 ATI-ia의 상용센서와 capacitve sensor를 사용하는 Robotus의 센서와 비교 하였으며, 비교하기 위해서 10초간 std deviation을 MINI-85와 Proposed Sensor의 각 축마다 계산을 하였으며, 이를 이용하여 resolution을 구할 수 있다. std deviation결과는 Table.~\ref{sensorstd}에 나와있다. 

In addition, a resolution evaluation was conducted to further assess the performance of the proposed sensor. The resolution was compared against two widely used commercial sensors: a strain-gauge-based sensor (MINI-85 by ATI Industrial Automation~\cite{ati_website}), which shares a similar elastomer structure, and a capacitive sensor developed by AIDIN ROBOTICS~\cite{aidinrobotics_website}. 

To quantify resolution, the standard deviation of each axis was calculated over a 10-second interval for both the MINI-85 and the proposed sensor. The computed standard deviations were then used to determine the resolution for each axis. The detailed results are summarized in Table~\ref{sensorstd}.

\begin{table}[!h]
\centering
\caption{Comparison of Standard Deviation Between Commercial Sensor and Proposed Sensor}
\begin{tabular}{ccccc}
\hline\hline
        & $F_x$\&$F_y$      & $F_z$      & $T_x$\&$T_y$      & $T_z$      \\
\hline 
Company(A)     & 0.116   & 0.175  & 0.00570  & 0.00332  \\
Proposed Sensor      & 0.0120  & 0.0386  & 0.00151  & 0.000574 \\
\hline
\end{tabular}
\label{sensorstd}
\end{table}
%이를 이용하면 proposed sensor의 resolution을 구할 수 있으며, 이 결과는 Table.~\ref{sensorresolution}에 나타냈으며 결과는 resolution (N, N$\cdot$m)과 Quantization Levels로 나타냈다.  
Using this method, the resolution of the proposed sensor was determined, and the results are presented in Table~\ref{sensorresolution}. The outcomes are reported in terms of resolution (in N and N$\cdot$m) and the corresponding number of quantization levels.

\begin{table}[!h]
\centering
\caption{Comparison of Sensor Resolution Between Commercial Sensors and Proposed Sensor}
\renewcommand{\arraystretch}{1.2}  % 행 간격 조정 (선택사항)
\begin{tabularx}{\linewidth}{l *{4}{>{\centering\arraybackslash}X}}
\hline\hline
Resolution       & $F_x$\&$F_y$ (N) & $F_z$ (N) & $T_x$\&$T_y$ (N$\cdot$m) & $T_z$ (N$\cdot$m) \\
\hline 
Company(A)       & 0.321   & 0.429   & 0.0134  & 0.00936 \\
Company(B)       & 0.150     & 0.150     & 0.0150    & 0.0150 \\
Proposed Sensor  & \textbf{0.0347}  & \textbf{0.0945}   & \textbf{0.00365} & \textbf{0.00163} \\
\hline
Quantization Levels & $F_x$\&$F_y$ & $F_z$ & $T_x$\&$T_y$ & $T_z$ \\
\hline
Company(A)       & 11822    & 17733   & 11968    & 17097 \\
Company(B)       & 4000     & 4000    & 3333     & 3333 \\
Proposed Sensor  & \textbf{51312}    & \textbf{30370}   & \textbf{14810}    & \textbf{55351} \\
\hline
\end{tabularx}
\label{sensorresolution}
\end{table}

As shown in Table~\ref{sensorresolution}, the proposed sensor demonstrates a resolution of approximately 0.035~N for the $x$- and $y$-axis forces, which is roughly nine times finer than that of the reference sensor. Additionally, the resolution in the $T_z$ torque is as low as 1.6~mN$\cdot$m, indicating high sensitivity in torque measurements as well.

The quantization level, defined as the ratio of the full-scale range to the resolution, serves as an effective indicator of how finely the sensor can distinguish force or torque inputs. The proposed sensor achieves up to 55,351 quantization steps, which is more than twice the granularity offered by conventional sensing methods.

%Table.\ref{sensorresolution}을 보면 센서의 resolution은 x,y축 힘은 0.035N정도로 9배정도 차이가 나며, z축 모멘트도 1.6mN$\cdot$m로 해상도가 좋은 것을 알 수 있다. 여기서 Quantization Level은 Full-Scale range에서 resolution을 나눈 값으로 얼마나 resolution이 더 좋은지를 판단하는 지표로 사용할 수 있는 것이다. 여기서 proposed sensor는 최대 55351 step으로 나눌 수 있어 기존의 방식보다 전체적으로 2배 이상 좋은 것을 알 수 있다.  

\begin{table}[!h]
\centering
\caption{Crosstalk Percentage Error Matrix (\%)}
\begin{tabular}{lcccccc}
\hline\hline
        & $F_x$     & $F_y$     & $F_z$     & $T_x$     & $T_y$     & $T_z$     \\
\hline
$F_x$ & 100. & 0.193 & 0.152 & 0.0248 & 0.153 & 0.0279 \\
$F_y$ & 0.0907 & 100. & 0.0773 & 0.0642 & 0.00533 & 0.0102 \\
$F_z$ & 0.236 & 0.128 & 100. & 0.0161 & 0.0146 & 0.0345 \\
$T_x$ & 0.175 & 0.301 & 0.195 & 100. & 0.00125 & 0.150 \\
$T_y$ & 0.596 & 0.719 & 0.385 & 0.152 & 100. & 0.122 \\
$T_z$ & 0.0654 & 0.201 & 0.0555 & 0.0457 & 0.00600 & 100. \\
\hline
\end{tabular}
\label{cross}
\end{table}

%크로스톡도 평가를 진행하였다. 각축에 힘을 주고 여러 축에 힘을 동시에 가하여 측정을 하였다. Crosstalk은 \%로 나타냈는데 이는 Full-scale percentage이며, 센서의 range인 Table.~\ref{sensrange}의 값을 이용하였다. 예를 들어서 x-axis force는 890~N의 힘이기 때문에 1780~N을 이용하여 계산을 한 값들이다. 
Crosstalk was also evaluated by applying force to individual axes as well as simultaneously to multiple axes. The crosstalk values are expressed as a percentage, representing the full-scale ratio based on the sensor ranges listed in Table~\ref{sensrange}. For example, the x-axis force has a range of 890~N, and therefore the full-scale reference value used for the calculation was 1780~N.

\section{Discussion and Conclusion}

\begin{figure}
    \centering
    \includegraphics[width=\linewidth]{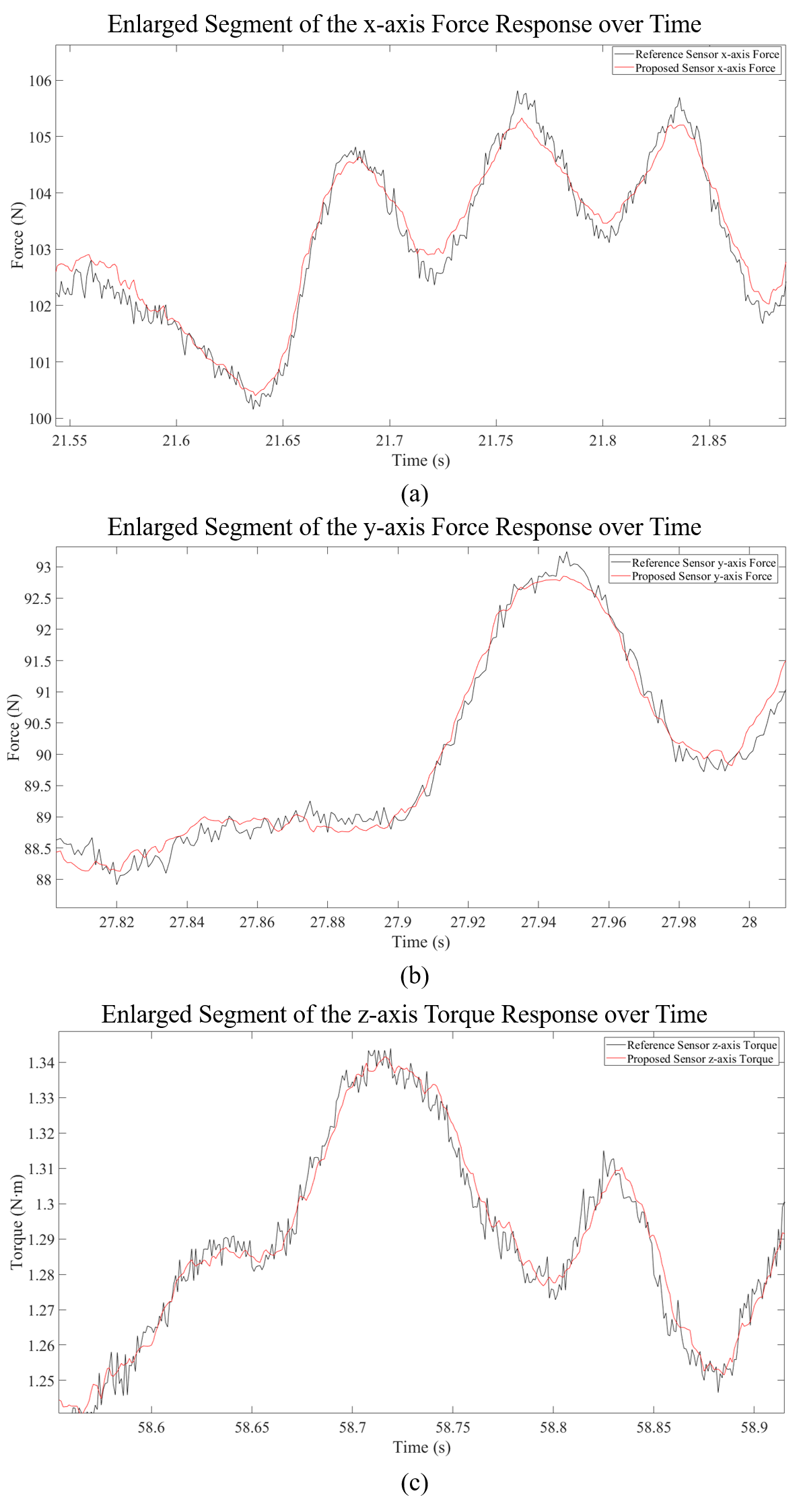}
    \caption{Magnified static F/T responses for evaluation: (a) Force along $x$-axis ($F_x$); (b) Force along $y$-axis ($F_y$); (c) Torque about $z$-axis ($M_z$). }
    \label{whak}
\end{figure}
The primary contribution of this work lies in the novel inductive sensing structure, mechanical integration, and system-level validation. While nonlinear fitting techniques were used to improve calibration accuracy, the focus of the study is not on the calibration algorithm itself but on demonstrating the sensor’s overall performance and practical applicability.

The proposed calibration process was validated through extensive static and repeatability experiments, showing consistent results with low RMSE and percentage error across all axes. These results indicate that the system operates robustly within the expected sensing range, even without explicit noise sensitivity analysis.

Among the evaluated fitting models, the rational function achieved the lowest RMSE despite using only five parameters, whereas other models, such as the Gaussian mixture and sigmoid models, employed six parameters. This suggests that the rational model provides a more efficient and generalizable representation of the underlying nonlinearity. Therefore, the choice of rational fitting is justified not only by accuracy but also by model simplicity and robustness against overfitting.

Fig.~\ref{whak} presents an enlarged view of the static force response. Fig.~\ref{whak}(a) shows the x-axis force, (b) shows the y-axis force, and (c) depicts the z-axis torque. In each plot, the red line represents the proposed sensor, while the black line corresponds to the reference sensor. A comparison of the two responses reveals that the proposed sensor exhibits significantly lower noise levels.

In terms of performance, the proposed sensor demonstrates superior resolution, achieving up to four times higher quantization levels compared to a commercial sensor. Most commercial sensors rely on external DAQ units and, in the case of strain-gauge-based systems, require both a Wheatstone bridge and an amplifier to detect strain variations. This amplification process inherently introduces noise, and achieving high resolution typically necessitates extensive filtering.

In contrast, the proposed sensor does not require amplification, as it measures inductance variation directly. This structural advantage inherently results in lower noise. Furthermore, the system supports a maximum sampling rate of approximately 4~kHz, which is sufficiently fast for dynamic measurements. Many commercial sensors operate at sampling rates around 200~Hz, indicating that the proposed sensor also offers benefits in terms of data acquisition speed. Additionally, due to its non-contact measurement principle, the proposed sensor provides improved robustness against impact and long-duration loading, unlike traditional contact-based strain-gauge sensors.

\begin{table*}[!t]
\begin{center}
\caption{Comparison with other F/T sensors}
\label{tab11}

\begin{tabular}{c c c c c c c c c c}
\hline\hline
  & Accuracy(FSE(\%)) &Sampling Rate&Resolution &Contact\\

 \hline
 \cellcolor[HTML]{EFEFEF} \bf{Proposed Method } &\cellcolor[HTML]{EFEFEF}1.0 &\cellcolor[HTML]{EFEFEF}$\sim$4,000Hz&\cellcolor[HTML]{EFEFEF}15-bit w/o external DAQ&\cellcolor[HTML]{EFEFEF}Non-contact\\

 Optical type~\cite{kim2024compact}&2.0&5,000~Hz&14-bit w/o external DAQ&Non-contact\\
Capacitor type~\cite{kim2016novel}&4.1&200~Hz&11-bit w/o external DAQ&Non-contact\\
Magnetic type~\cite{ananthanarayanan2012compact}&22.5&2~ms delay&low&Non-contact\\
Pressure Sensing type~\cite{tar2011development}&22.2&1,000~Hz&Low&Contact\\
  RFT80-6A01(Commercial)~\cite{robotous_website} &3.0  &1,000~Hz&13-bit w/o external DAQ&Non-contact\\
AFT200-KIT(Commercial)~\cite{aidinrobotics_website} &-  &1,000~Hz&11-bit w/o external DAQ&Non-contact\\
  SensOne-Serial(Commercial)~\cite{botasys_website} &2.0  &800~Hz&11-bit w/o external DAQ&Contact\\
  MINI-85(Commercial)~\cite{ati_website} &$\sim$ 2.0  &7,000~Hz&13-bit with DAQ&Contact\\
    \hline
   & Weight&Load Capacity &Number of Sensors &Sensing Method\\
  \hline
 \cellcolor[HTML]{EFEFEF} \bf{Proposed Method } &\cellcolor[HTML]{EFEFEF}280g&\cellcolor[HTML]{EFEFEF}2,870~N &\cellcolor[HTML]{EFEFEF}6 &\cellcolor[HTML]{EFEFEF} Inductive Sensor \\
Optical type~\cite{kim2024compact}&79~g&3,700~N&6&Photocoupler\\
Capacitor type~\cite{kim2016novel}&15~g&100~N&6&Capacitive Sensor\\
Magnetic type~\cite{ananthanarayanan2012compact}&-&1,000~N&5(2DOF)&Magnetic Sensor\\
Pressure Sensing type~\cite{tar2011development}&-&450~N$\sim$&9(3DOF)&Pressure Sensor\\
  RFT80-6A01(Commercial)~\cite{robotous_website} &226~g&800~N&6&Capacitive Sensor \\
  AFT200-KIT(Commercial)~\cite{aidinrobotics_website} &236~g  &400~N&6&Capacitive Sensor\\
  SensOne-Serial(Commercial)~\cite{botasys_website}& 235~g&2,400~N&12~&Strain-gauge \\
   MINI-85(Commercial)~\cite{ati_website}& 635~g&3,800~N&12~&Strain-gauge \\

  \hline
\end{tabular}

\end{center}
\end{table*}

% Table~\ref{tab11}을 보면 proposed sensor와 다른 센서연구와, 상용 센서들 간의 비교를 나타낸 것이다. 무게는 accuracy도 magnetic과 pressure type보다 우수하면서 sampling rate는 Proposed sensor가 4~kHz로 일반적인 상용센서와 다르게 빠른 편에 속하면서, resolution이 제일 높은 것을 확인 할 수 있다. proposed sensor는 inductive sensing을 기반으로 noncontact 방식을 사용하였으며, 센서 개수도 6개로 적은 것을 볼 수 있다. Photocoupler를 이용하여 제작한 센서나 Capacitive sensor를 이용한 센서는 소형화에 특화되어있으며, Strain-gauge를 사용한 센서들은 보통 크기가 크며, 6축을 측정하기 위해서 최소 12개의 strain-gauge를 부착을 해야한다. magnetic sensor나 pressure sensor는 resolution이 낮으며, 정확도가 안좋은 것을 볼 수 있다. Proposed sensor는 지금 Inductive sensing기반으로 측정을 하지만, 여기서 coil turn수나 layer 선정 또는 conductive target과의 거리가 최적화가 안되어있기 때문에 이것이 이 연구의 한계점이다. future work로써 coil turn수와 layer 갯수와 conductive target의 거리를 최적화하게 되면 지금보다 더 좋은 성능을 보일 것으로 전망된다. 또한, 온도에 따른 보상도 같이 하여 robust한 센서를 만들고자 한다. 
Table~\ref{tab11} provides a comparison between the proposed sensor, other research-based sensors, and commercially available six-axis force/torque sensors. The proposed sensor demonstrates superior accuracy compared to magnetic and pressure-based sensors while also achieving the highest resolution among all compared systems. Moreover, with a sampling rate of 4~kHz, it operates significantly faster than typical commercial sensors.

The proposed system supports a maximum sampling rate of approximately 4~kHz, which is the upper limit defined by the LDC1614 hardware specifications. However, in the static and comparative experiments conducted in this study, a sampling rate of 1~kHz was employed to ensure compatibility with the reference sensor and maintain signal stability.

The proposed sensor is based on inductive sensing and utilizes a non-contact measurement approach. It employs only six sensing elements, which is relatively low compared to other configurations. Sensors developed using photocouplers or capacitive principles are generally optimized for miniaturization. In contrast, strain-gauge-based sensors tend to be bulky and require a minimum of 12 strain-gauges to measure six degrees of freedom. Magnetic and pressure-based sensors typically suffer from low resolution and poor accuracy. The system's compact design, non-contact operation, and integrated high-speed signal processing unit highlight its suitability for embedded robotic applications.

While the proposed sensor currently employs inductive sensing, its performance is still limited by non-optimized parameters such as the number of coil turns, the number of layers, and the distance to the conductive target. These factors represent the current limitations of this study. For example, the horizontal sensing coils-which are responsible for measuring the $x$- and $y$-axis forces and the $z$-axis torque-currently exhibit higher resolution than the vertical sensing coils, which measure the $x$- and $y$-axis torques and the $z$-axis force. This discrepancy suggests that further optimization of the coil turns and the distance to the target could lead to improved performance in all measurement axes. 

As future work, performance is expected to improve through optimization of the coil turns, layer configuration, and spacing to the conductive target. Additionally, temperature compensation will be incorporated to further enhance sensor robustness under varying environmental conditions. In addition, further investigation is required to assess the sensor's robustness in electromagnetic interference (EMI) and electromagnetic compatibility (EMC) environments. Due to the nature of inductive sensing, the system may be susceptible to electromagnetic disturbances. Therefore, comprehensive stability testing and the implementation of shielding strategies in the design should be considered to ensure reliable operation under such conditions. While a formal sensitivity analysis of the nonlinear fitting function is not included, the system demonstrated reliable performance under various test conditions. More detailed noise analysis could be a topic for future exploration if the calibration method is further generalized or applied to different sensor platforms.

\bibliographystyle{Bibliography/IEEEtranTIE}
% \bibliography{Bibliography/IEEEabrv,forceref} %IEEEabrv instead of IEEEfull

% \vspace{-1cm}
\begin{IEEEbiography}[{\includegraphics[width=1in,height=1.25in,clip,keepaspectratio]{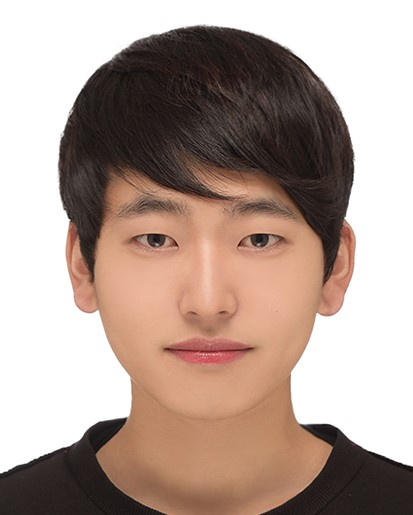}}]
{Hyun-Bin Kim}
~ obtained his B.S., M.S., and Ph.D. degrees in Mechanical Engineering from Korea Advanced Institute of Science and Technology(KAIST), Daejeon, Republic of Korea, in 2020, 2022, and 2025 respectively. Currently serving as a post-doctoral researcher at KAIST, his current research interests include force/torque sensors, legged robot control, robot design, and mechatronics systems.
\end{IEEEbiography}

\begin{IEEEbiography}[{\includegraphics[width=1in,height=1.25in,clip,keepaspectratio]{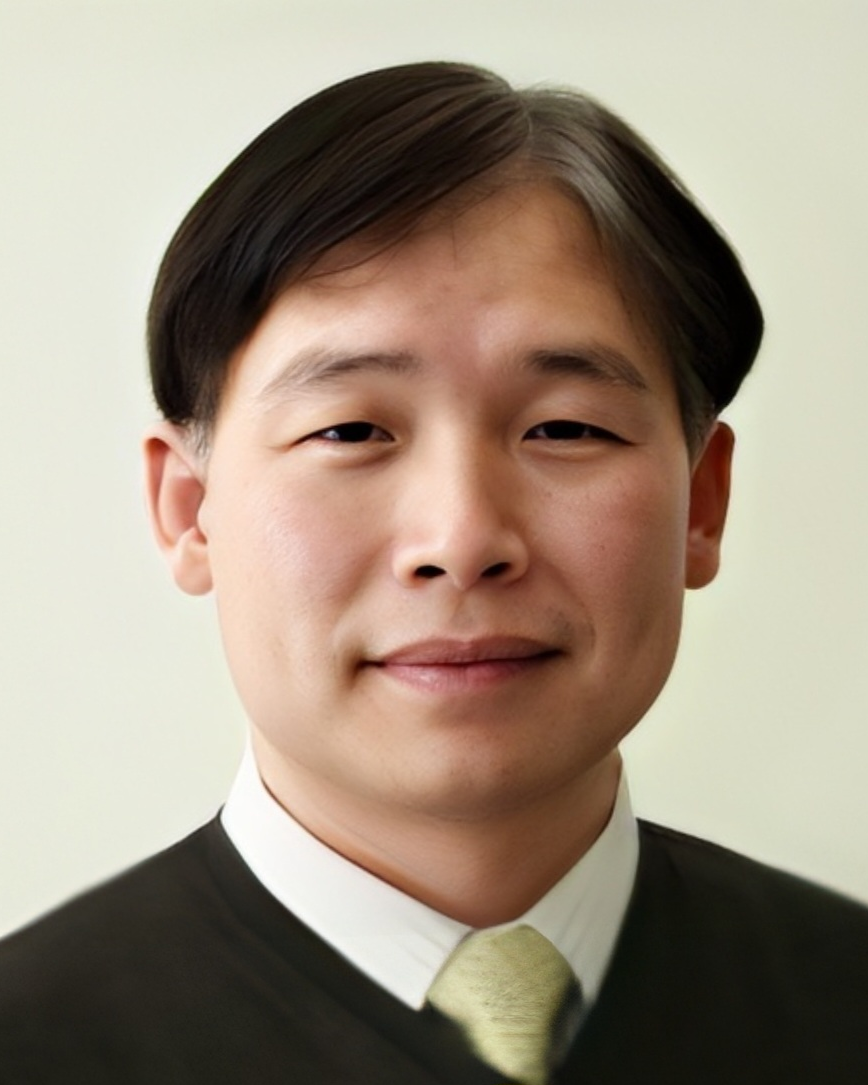}}]{Kyung-Soo Kim}~(Fellow, IEEE)~ obtained his B.S., M.S., and Ph.D. degrees in Mechanical Engineering from Korea Advanced Institute of Science and Technology (KAIST), Daejeon, Republic of Korea, in 1993, 1995, and 1999, respectively. Since 2007, he has been with the Department of Mechanical Engineering, KAIST. 
His research interests include control theory, electric vehicles, and autonomous vehicles. He serves as an Associate Editor for the Automatica and the Journal of Mechanical Science and Technology.
\end{IEEEbiography}

\end{document}